\documentclass[aps,prb,tightenlines,twocolumn,floatfix,superscriptaddress]{revtex4}
\usepackage{graphicx}
\usepackage{amssymb,amsthm,bbm}

\usepackage{algorithm}
\usepackage{algpseudocode}
\usepackage{amsmath}
\usepackage{amsfonts,ragged2e}
\usepackage{dsfont}
\usepackage{color}
\usepackage{braket}
\graphicspath{{../Figures/}}
\usepackage[singlelinecheck=off]{subcaption}
\usepackage[singlelinecheck=off, justification=raggedright]{caption}

\begin{document}

\title{Mode-Assisted Joint Training of Deep Boltzmann Machines}

\author{Haik Manukian}
\email{email: hmanukia@ucsd.edu}
\affiliation{Department of Physics, University of California, San Diego, La Jolla, CA 92093}

\author{Massimiliano Di Ventra}
\email{email: diventra@physics.ucsd.edu}
\affiliation{Department of Physics, University of California, San Diego, La Jolla, CA 92093}

\begin{abstract}
The deep extension of the restricted Boltzmann machine (RBM), known as the deep Boltzmann machine (DBM), is an expressive family of machine learning models which can serve as compact representations of complex probability distributions. 
However, {\it jointly} training DBMs in the {\it unsupervised} setting has proven to be a formidable task. 
A recent technique we have proposed, called mode-assisted training, has shown great success in improving the unsupervised training of RBMs. 
Here, we show that the performance gains of the mode-assisted training are even more dramatic for DBMs. 
In fact, DBMs jointly trained with the mode-assisted algorithm can represent the same data set with {\it orders of magnitude} lower number of total parameters compared to state-of-the-art training procedures and even with respect to RBMs, provided a {\it fan-in} network topology is also 
introduced. 
This substantial saving in number of parameters makes this training method very appealing also for hardware implementations. 
\end{abstract}

\maketitle

One of the most highly influential models in Artificial Intelligence (AI) and deep learning in particular is the Boltzmann machine (BM). 
They were constructed\cite{ackley1985learning,smolensky1986information} as a powerful stochastic generalization of Hopfield networks~\cite{hopfield1982neural} that possess a simple expression for their log-likelihood gradient.
However, they remained impractical to train due to their reliance on high-dimensional sampling to calculate that gradient.
A relatively efficient learning algorithm, called contrastive divergence (CD), was discovered for BMs with a simplified topology called restricted Boltzmann machines (RBMs)\cite{hinton2002training}, which have since gone on to see success in various domains~\cite{goodfellow2016deep}. 
However, the extension of RBMs to deep Boltzmann machines (DBMs) has been difficult~\cite{goodfellow2012scaling}, and as such, DBMs are now mostly overshadowed by their deep feedforward cousins~\cite{goodfellow2014generative} in generative applications.

This is not due to DBM's lack of ability, but rather the absence of effective means to train these models.
There remain quite a few reasons to search for better learning algorithms for DBMs, as they are a versatile computational medium. 
A principle use is as compact generative models for complex probability distributions in unsupervised settings, considered a critical component of the forthcoming ``third-wave'' of AI~\cite{launchbury2017darpa}. 
Trained DBMs can also serve as an informed prior for feedforward networks, leading to better generalization in supervised tasks~\cite{erhan2010does}. 
In the physical sciences, DBMs serve as powerful variational representations of many body wavefunctions, more efficiently~\cite{gao2017efficient} than RBMs~\cite{carleo2017solving,melko2019restricted}, and have potential applications in condensed matter physics and quantum computing~\cite{carleo2019machine}.

\begin{figure}[t]
\centering
		\includegraphics[width=0.4\textwidth]{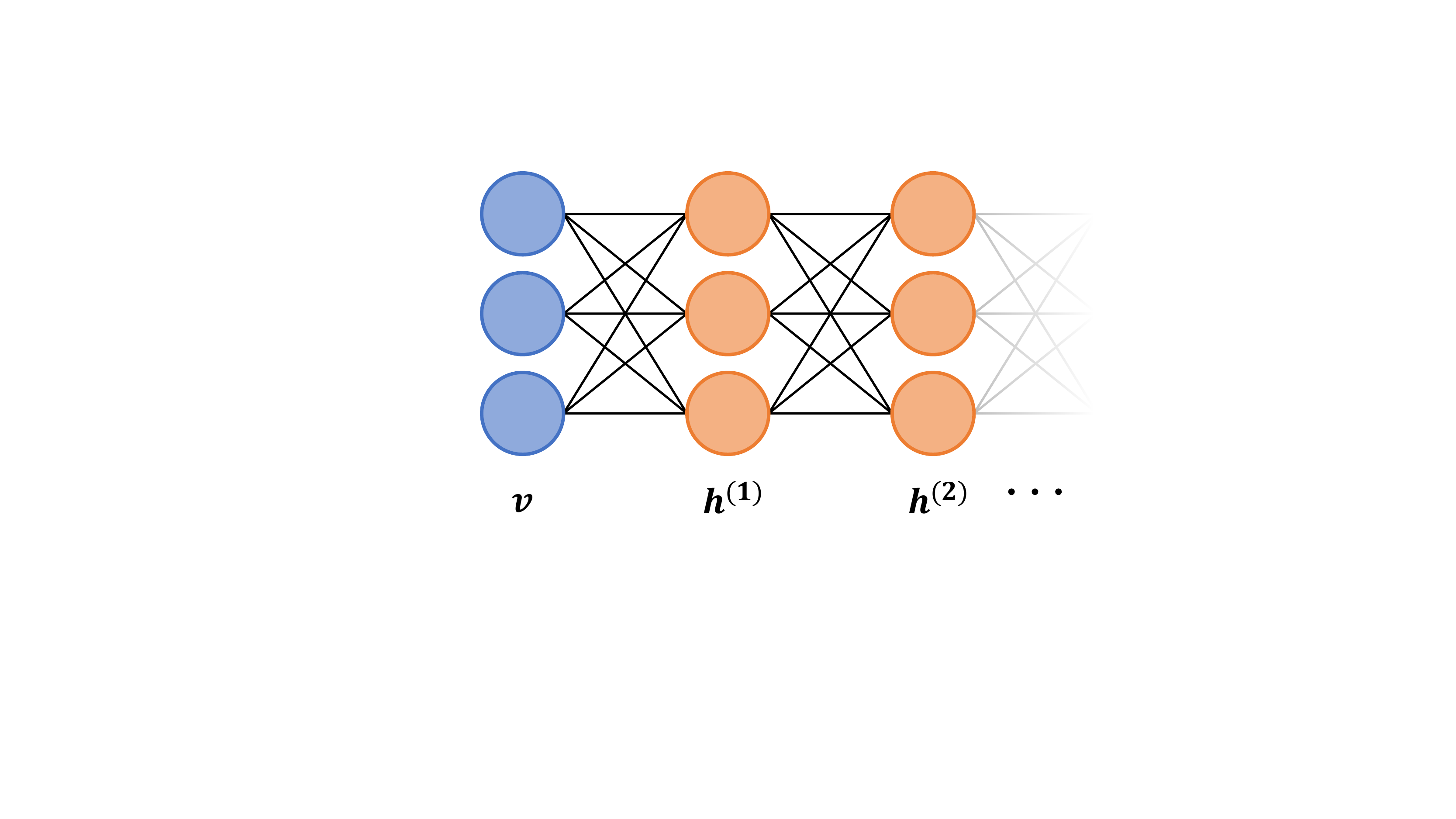}	
	\caption{Schematic of a deep Boltzmann Machine with a visible layer, $v$, and hidden layers, $h^{(j)}$, with $j=1, 2, \dots$. Connections between nodes are symmetric and undirected, in contrast to typical directed feedforward networks.}
	\label{fig:dbm}
\end{figure}

Various attempts have been made to climb the summit of DBM training~\cite{salakhutdinov2009deep, salakhutdinov2010efficient, hinton2012better,goodfellow2013joint,melchior2016center}. 
Most approaches rely on pre-training by breaking up layers into RBMs and training them sequentially, after which the DBM is fine-tuned jointly. 
Correlations between layers are ignored during pre-training, minimizing the potential advantages of the deep architecture and can be disruptive to joint training~\cite{melchior2016center}. 
Recently, the authors introduced mode-assisted training~\cite{AcceleratingDL,manukian2020mode}, which combines CD with samples of the model distribution mode. 
This stabilizes training, allows the learning of very accurate densities, and strikes a better tradeoff between accuracy and computational cost compared to CD. 
As the method is agnostic to the connectivity of the network, we can apply mode-assistance to DBMs with the hope of capturing the model capacity missed by other approaches.

In this work, we find that the benefits of mode-assisted training are even more dramatic in the case of DBMs. In fact, it produces more accurate models without requiring pre-training while also utilizing {\it orders of magnitude} less parameters, compared to pre-trained~\cite{salakhutdinov2009deep} or centered DBMs~\cite{melchior2016center}. 
The role of the network topology is also discussed, where we discover that DBMs are easier to train if the size of the layers decreases with depth.
We evaluate the density modeling performance of mode-assisted DBMs by computing exact log-likelihoods achieved on small data sets and approximating the likelihood on the MNIST data set. The approach we propose can be extended to other types of neural networks, and 
is relevant also for the hardware implementation of these models, where a much smaller number of parameters directly translates into 
components and energy savings.

To see how mode-assisted training can be extended to deep architectures, we give a quick overview of DBMs and the basic approach to their training. 
DBMs are undirected weighted graphs that differentiate between $n_v$ visible nodes, and $\ell$ layers of $n_\ell$ latent, or `hidden', nodes, not directly constrained by the data~\cite{salakhutdinov2009deep}. 
We assume that $\ell >1$ as $\ell = 1$ recovers the RBM, and like an RBM, there are no connections within a layer. 
Each state of the machine corresponds to an energy of the form
\begin{align}
\label{eq:E}
E({\bf v}, {\bf h}^{(1)}, &\cdots , {\bf h}^{(\ell)}) =- {\bf a}^T{\bf v} - {\bf v}^T {\bf W}^{(1)} {\bf h}^{(1)} \\ \nonumber
&-\sum_{i=1}^\ell{\bf b}^{(i)T}{\bf h}^{(i)}- \sum_{i=2}^\ell {\bf h}^{(i-1)T}{\bf W}^{(i-1)}{\bf h}^{(i)},
\end{align}
where the biases ${\bf a} \in \mathbb{R}^{n_0}$, ${\bf b}^{\ell} \in \mathbb{R}^{n_\ell}$, and weights ${\bf W}^\ell \in \mathbb{R}^{n_{\ell-1}\times n_{\ell}}$ are the learnable parameters.  
The energy function in Eq.~(\ref{eq:E}) induces a Boltzmann-Gibbs distribution over states,
\begin{equation}
\label{eq:pvh}
p({\bf x}) = \frac{e^{-E({\bf x})}}{\mathcal{Z}},
\end{equation}
where ${\bf x} = ({\bf v}, {\bf h}^{(1)},\dots,{\bf h}^{(\ell)})$. The partition function, $\mathcal{Z} = \sum_{\{{\bf x}\}} e^{-E({\bf x})}$, involves the sum of an exponentially growing number of states, making the exact computation of its value infeasible for most data sets. 

During learning, a DBM is tasked to match its marginal distribution over the visible layer,  $p({\bf v}) = \sum_{\{ {\bf h} \}} p({\bf v}, {\bf h})$, to an unknown data distribution, $q({\bf v})$, represented by a data set, $\mathcal{D}$. 
Training a DBM amounts to a search for the appropriate weights and biases that will minimize the quantity known as the Kullback-Leibler (KL) divergence between the two distributions,
\begin{equation}
\label{eq:kl}
\text{KL}(q || p) = \sum_{\{{\bf v}\}} q({\bf v}) \log \frac{q({\bf v})}{p({\bf v})},
\end{equation}
or, equivalently, maximizing the log-likelihood of the dataset, $\textrm{LL}(p) = \sum_{{\bf v}\in \mathcal{D}}\log p({\bf v})$. 
The optimization of the non-linear, and typically high dimensional Eq.~(\ref{eq:kl}) (or log-likelihood), is often done via stochastic gradient descent with respect to the DBM parameters, which leads to weight updates of the form~\cite{fischer2012introduction},
\begin{equation}
\label{eq:w_gradient}
\Delta w_{ij} \propto  \langle x_i x_j \rangle_{D} - \langle x_i x_j \rangle_{M}.
\end{equation}

For every gradient update in Eq.~(\ref{eq:w_gradient}), nodal statistics must be computed under two different distributions. 
The first one on the RHS of Eq.~(\ref{eq:w_gradient}) is called the ``data term'', and is an expectation over the data induced distribution, $q({\bf v})p({\bf h}| {\bf v})$, with the network's visible layer fixed to the data. 
The second  term on the RHS of Eq.~(\ref{eq:w_gradient}) is called the ``model term'' which is an expectation over the entire model distribution in Eq. (\ref{eq:pvh}). 
In the case of RBMs, the data term can be sampled from exactly, but the model term must be approximated. 
With DBMs, the data term must also be approximated, most popularly with an iterative mean field procedure (see Methods).

In both cases, model statistics are collected via a Markov Chain Monte Carlo (MCMC) procedure dubbed `contrastive divergence' (CD)~\cite{hinton2002training}. 
CD-$k$ is a form of Gibbs sampling that initializes chains of length $k$ from elements of the dataset. 
Trouble arises when the model distribution contains 'spurious' modes where the data distribution has negligible probability. 
In these cases, ergodicity breaks down, and mixing times become prohibitively long, frequently resulting in CD becoming biased enough to cause training to diverge~\cite{fischer2010empirical}. 
Training a DBM {\it jointly} with CD has proven to be a formidable task. 
Even a two-layer DBM on MNIST has not seen success without some kind of modification.~\cite{salakhutdinov2009deep,melchior2016center,goodfellow2012scaling,desjardins2012training}

Here, instead, we use mode-assisted training in the {\it joint} and {\it unsupervised} learning of DBMs. The essence of mode-assisted training is the replacement of some gradient updates in Eq.~\ref{eq:w_gradient} with ones of the form,

\begin{equation}
\Delta w_{ij} \propto [x_i x_j]_{D} - [x_ix_j]_{M}
\end{equation}

Where the notation $[f({\bf x})]_q$ represents $f({\bf x}_{\text{mode}})$, evaluated at ${\bf x}_{\text{mode}}$, the mode of some distribution, $q({\bf x})$. One may employ any optimization solver to sample the mode of the above distributions. Due to its proven 
efficiency, here we employ a memcomputing one as reported in our previous work~\cite{manukian2020mode}.

The mode-assisted update can be thought of as a saddle-point approximation of an expectation~\cite{bender2013advanced}. This technique, also known as Laplace's method, is commonly employed to approximate integrals (expectations) of the form,

\begin{equation}
\langle f({\bf x}) \rangle = \frac{\int e^{-E({\bf x})}f({\bf x})d{\bf x}}{\int e^{-E({\bf x})} d{\bf x}} \approx f({\bf x}_\text{mode}).
\end{equation}

\begin{figure*}[t!]
\centering
{\includegraphics[width=0.45\textwidth]{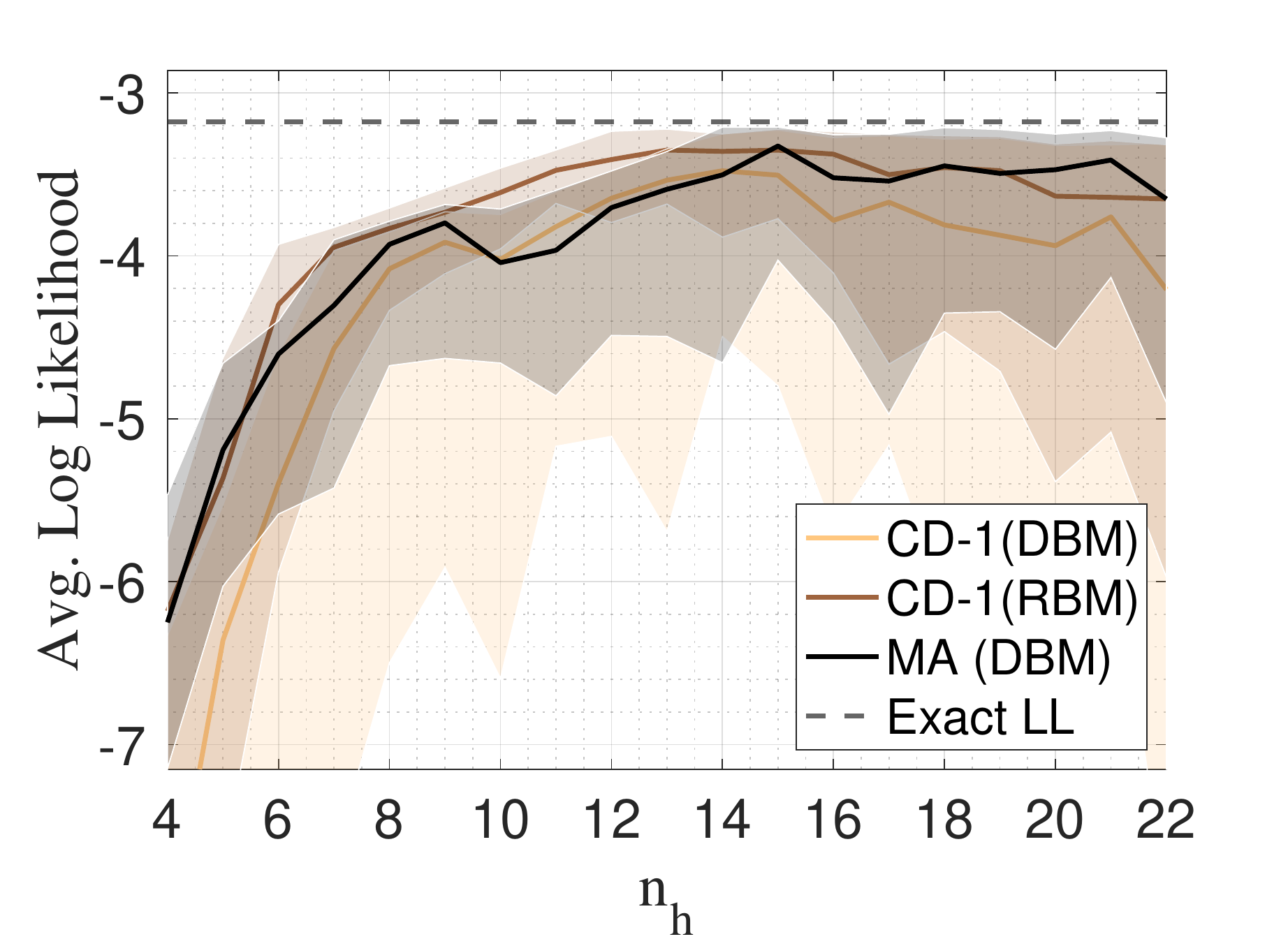}}%
\quad
{\includegraphics[width=0.45\textwidth]{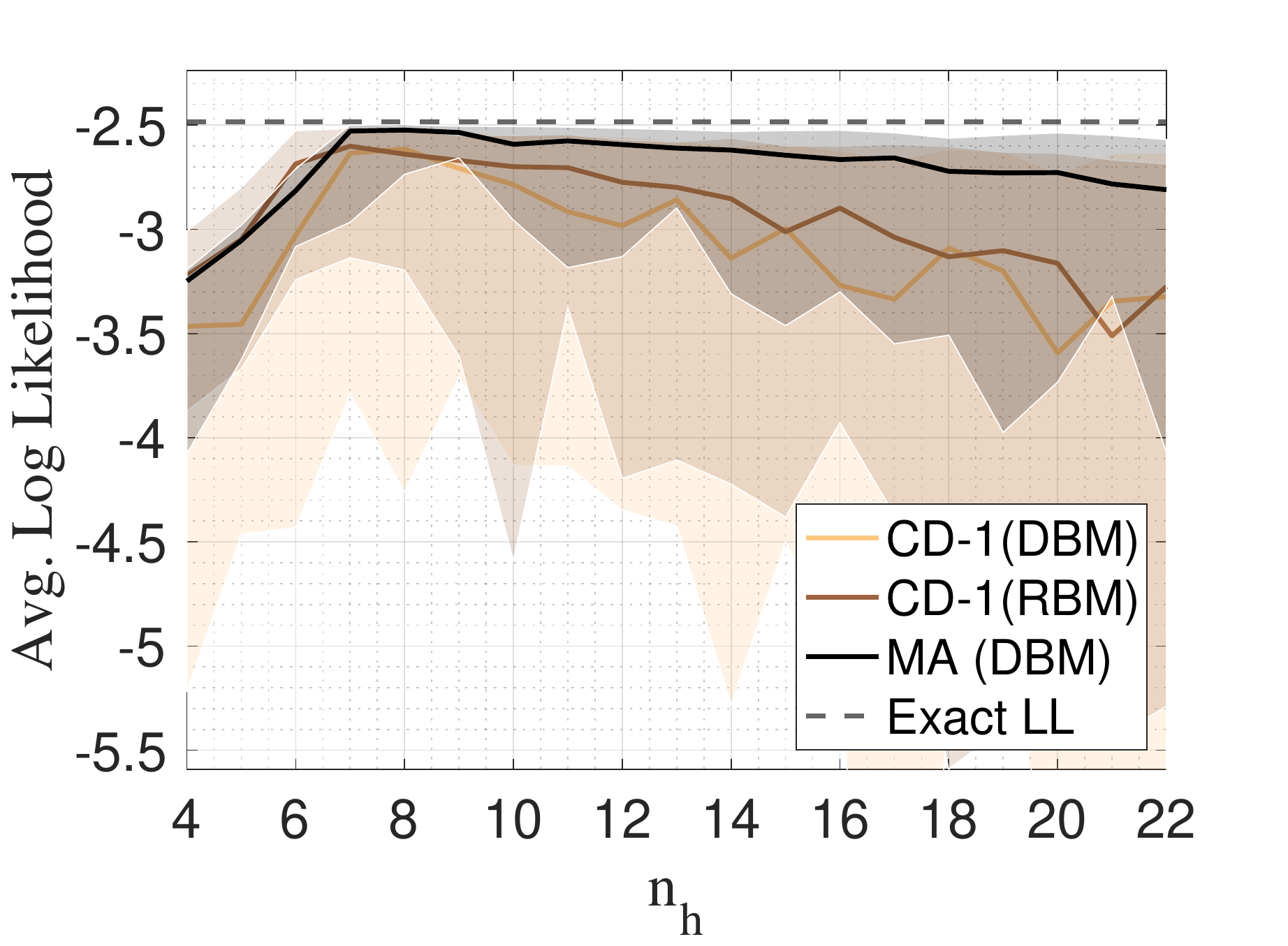}}%
	\caption{Average converged log-likelihood performance (lower bound) between RBMs trained with CD-1, mode-assisted DBMs (MA), and unassisted DBMs with CD-1. The 
		DBMs have two hidden layers. The networks were trained on the shifting bar data set with $n_v = 24$ (left plot) and $n_v = 12$ (right plot) for 200,000 and 100,000 gradient updates respectively, following a linearly decaying learning rate schedule from $\epsilon = 1 \to 0.001$. For the DBMs the hidden layer ratio was fixed at $\alpha = n_{h^{(2)}}/n_{h^{(1)}}=0.2$. Performance is shown as a function of total number of hidden nodes, $n_h=n_{h^{(1)}} + n_{h^{(2)}}$. The solid lines are the median obtained across an ensemble of 50 networks, and the shaded regions enclose the 95th and 5th percentiles.}
	\label{fig2}
\end{figure*}
The weight updates driven by the mode are incorporated in a probabilistic way, with the probability of a mode driven update following a sigmoid, starting low in the initial phases of the training and reaching a maximal value at the end of training: 

\begin{equation}
P_\text{mode}(n) = P_\text{max}\sigma(\alpha n + \beta).
\end{equation}

Here, $n$ is the current epoch, and $\alpha$, $\beta$, $P_\text{max}$ control the shape of the sigmoid. Throughout the work we set, $\alpha = 20/N$ ($N$ is the total number of epochs), $\beta = -6$ and $P_\text{max} = 0.1$.  This schedule was introduced in Ref.~\onlinecite{manukian2020mode}, based on the empirical observation that the mode samples work best when the support has been `discovered' by CD. 

The key insight of mode assisted training is {\it not} to approximate the expectation, but rather directly prevent spurious modes from appearing in the model distribution. This allows the inexact but efficient approximation of CD-$k$ to minimize the KL divergence (or maximize the log-likelihood) without diverging. Searching for the mode with a highly specialized optimizer results in a better trade-off between log-likelihood performance and computation time.


\section*{Results}
To illuminate the effectiveness of mode assistance on DBMs, we first evaluate performance on two differently sized synthetic shifting bar data sets. 
In Fig.~\ref{fig2}, the mode-assisted algorithm on a two-layer DBM is compared to two baselines, CD on the same DBM and to CD on an RBM with the same number of hidden nodes. 
The converged log-likelihoods are reported, and results are shown as a function of the total number of hidden nodes. 
Overall, mode-assisted training almost always converges to more accurate densities than CD alone on a DBM, and in most cases also does better than the corresponding RBM with CD. 
Mode assistance is also seen to prevent the divergence sometimes seen with Gibbs sampling, as well as reducing the variance of the converged models, resulting in smaller error bars. Past a certain point, all methods expectedly incur a loss in performance as the number of hidden nodes increases for a fixed number of training iterations. Mode-assisted training suffers the least in this regard.

Although the DBMs in Fig.~\ref{fig2} possess the same total number of hidden nodes as the RBMs, they contain fewer total parameters. This means that a better trained DBM takes advantage of abstract features composition afforded by the depth of the network, mirroring similar gains found in deep feedforward neural networks compared to single layer perceptrons~\cite{goodfellow2016deep}. 

Note, however, that this parameter efficiency in DBMs is {\it not} present when training jointly with CD. In fact, they perform systematically worse than an equivalently sized RBM in terms of log-likelihood. Mode assisted DBMs on the other hand perform as well as RBMs or better, all the while maintaining parameter efficiency.

To quantify this efficiency gain, let us consider the case of a DBM with two hidden layers. If the total number of hidden nodes is fixed, $n_h = n_{h^{(1)}} + n_{h^{(2)}}$, then a measure of parameter efficiency compared to an RBM with the same number of nodes is captured by the parameter $\alpha = n_{h^{(2)}}/n_{h^{(1)}}$. The total number of parameters (weights and biases) in an RBM with $n_v$ visible nodes and $n_h$ hidden nodes is,

\[
n_{\text{RBM}} = n_vn_h + n_v + n_h.
\]
A DBM with the same total number of hidden nodes would have the following number of parameters
\[
n_{\text{DBM}} = n_vn_{h^{(1)}} + n_{h^{(1)}}n_{h^{(2)}} + n_v + n_{h^{(1)}} + n_{h^{(2)}}.
\]

Considering typical training scenarios of a large visible layer, $n_v \gg n_h \gg 1$, the fraction of total parameters used in the DBM compared to the RBM (the {\it parameter efficiency}) can be simplified as,

\begin{equation}
e = \frac{n_{\text{DBM}}}{n_{\text{RBM}}}\approx \frac{1}{1 + \alpha}.
\label{eq:eff}
\end{equation}

\begin{figure}[h]
\centering
		\includegraphics[width=0.45\textwidth]{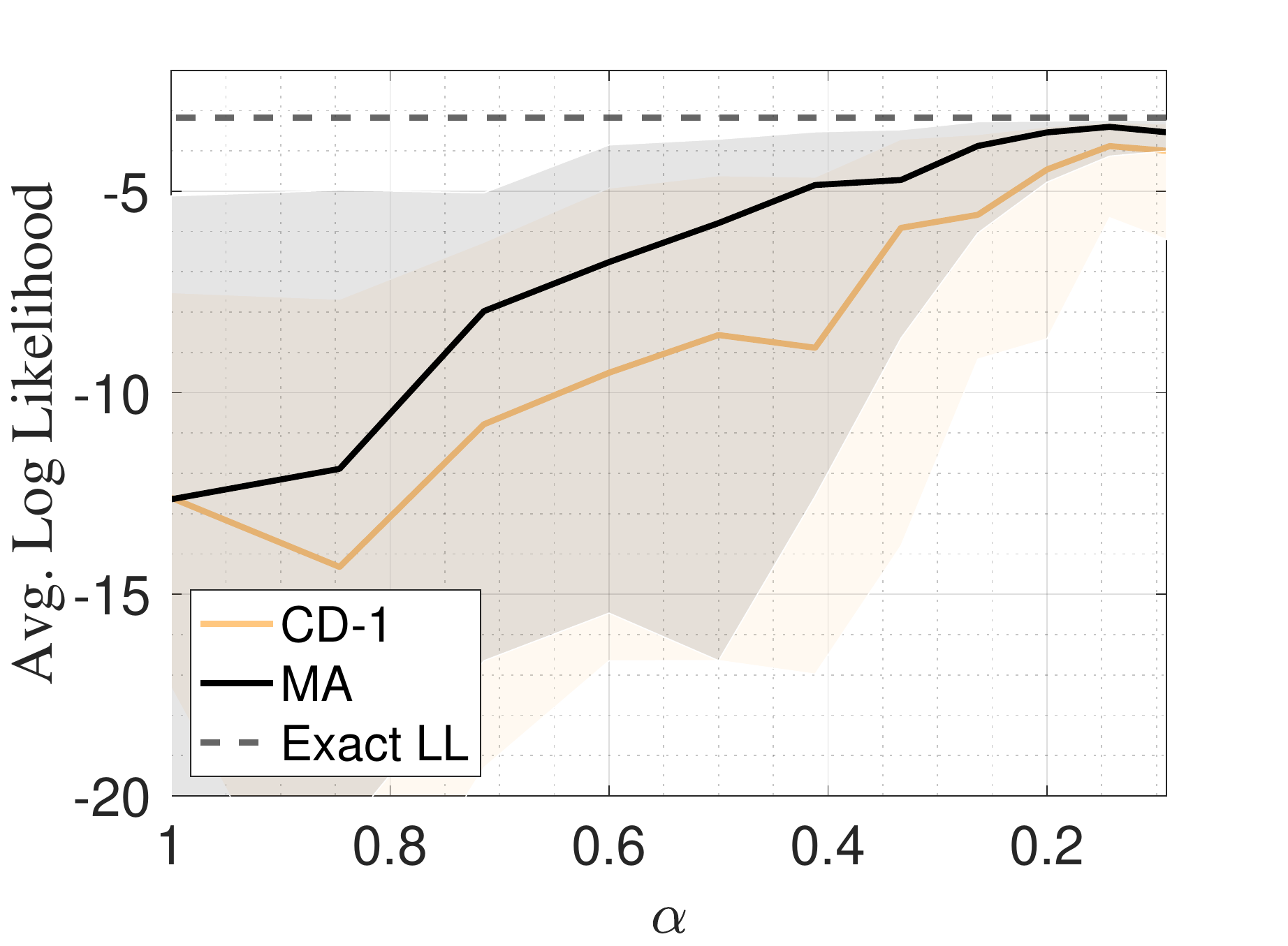}	
	\caption{Average Log-Likelihood achieved on an $n_v = 24$ dimensional shifting bar data set as a function of DBM shape, $\alpha$. The total number of hidden nodes are kept fixed at $n_{h^{(1)}} + n_{h^{(2)}} = 22$. An ensemble of 50 DBMs were trained with CD-1 and MA, where the solid line shows the median average log-likelihood achieved after $10^5$ gradient iterations with a linearly decaying learning rate $\epsilon = 1 \to 0.01$, and shaded regions delineate the 95th and 5th percentiles.}
	\label{fig:dbm_alpha}
\end{figure}

The smaller $e$, the fewer parameters a DBM possesses with respect to an RBM with the same number of nodes. 
We then see that a na\"ive interpretation of Eq.~(\ref{eq:eff}) would suggest that the DBM parameter efficiency becomes more significant with increasing $\alpha$, meaning the resulting DBM has more neurons in the deeper layers (fans out). However, this assumes equal log-likelihood (performance) of the two models, which is not obvious. In fact, for DBMs we expect that a {\it redistribution} of the relative 
number of nodes in the different hidden layers plays a significant role in their performance. This tradeoff between parameter efficiency and performance merits further investigation.

This is shown in Fig. \ref{fig:dbm_alpha}, where DBMs trained with and without mode assistance are compared as a function of $\alpha$. 
For all cases, the mode-assisted algorithm performs better than its unassisted counterpart, which is nothing other than the confirmation 
of the results of Fig.~\ref{fig2}. 
Importantly, however, we also observe that the log-likelihood performance {\it increases} as the network becomes less parameter efficient compared to an RBM $(\alpha \to 0)$, namely when it has a {\it fan-in} topology. 
We find that a balance is reached around $\alpha \sim 0.15$, namely the second hidden layer has only about 15\% of nodes than the first hidden layer. The message here is that for a {\it fixed} number of hidden nodes, it is best to organize the network as a fan-in DBM rather than an RBM, if one has a method to train the DBM close to capacity in the first place.

\begin{figure}[h]
\centering
		\includegraphics[width=0.45\textwidth]{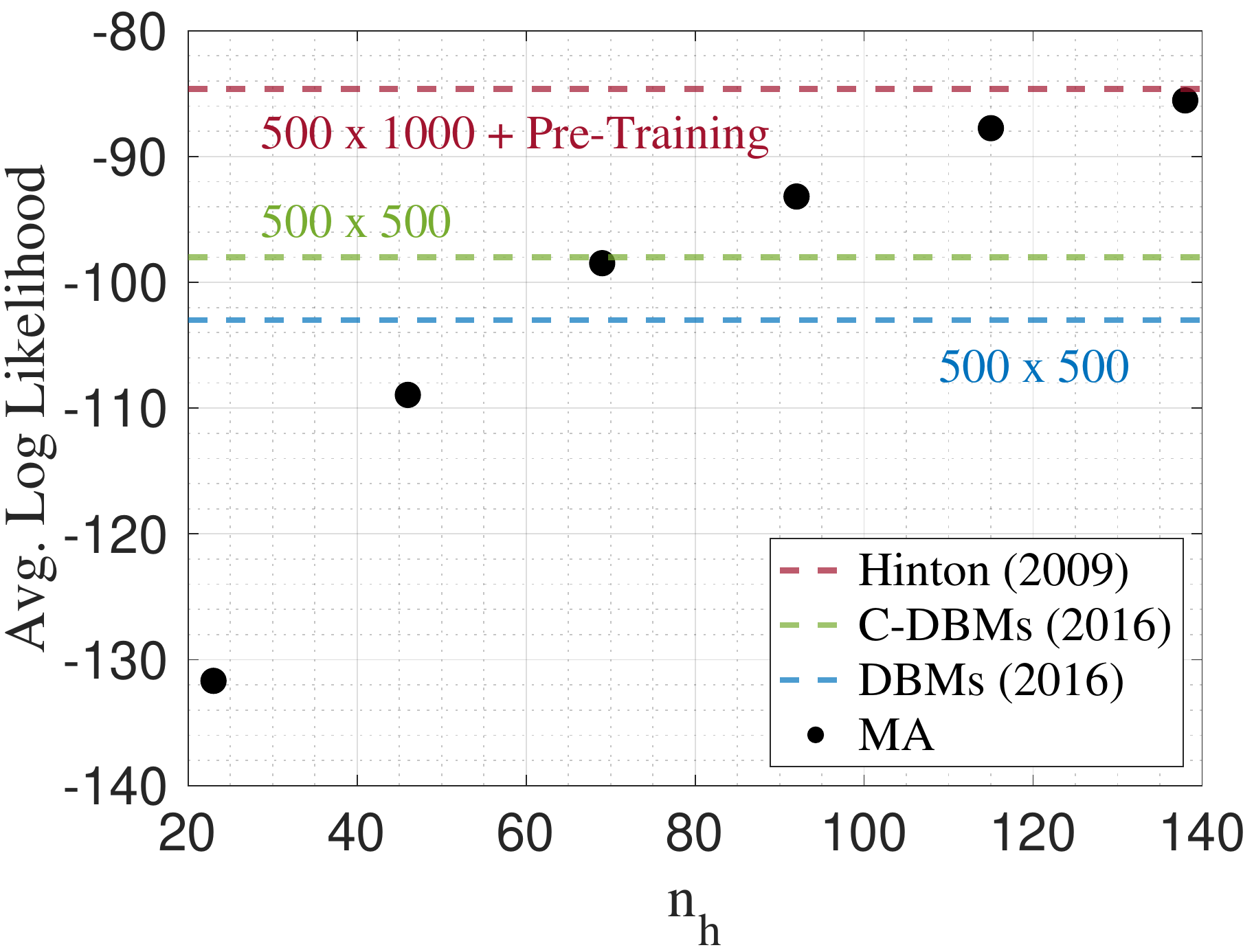}	
	\caption{Average log-likelihood on the MNIST data set achieved after 100 epochs with mode-assisted training (MA) compared to the recent best achieved results on DBMs using CD as well as pre-training. DBMs trained with MA were all with fixed $\alpha = 0.15$ and an increasing number of hidden nodes, $n_h$. The learning rate was chosen to follow a linear decay, $\epsilon = 0.05 \to 0.0005$ and no pre-training was employed. The best log-likelihood was reported out of 10 randomly initialized runs.}
	\label{fig:dbm_mnist}
\end{figure}

Finally, to show the substantial improvements provided by the mode-assisted training of DBMs, we compare it to the training of centered DBMs\cite{melchior2016center} and pre-trained DBMs\cite{salakhutdinov2009deep} using CD on the MNIST data set. In Fig.~\ref{fig:dbm_mnist}, we report similar trends we observed in the smaller data sets, but scaled to more dramatic results. As the number of hidden nodes increases, small mode-assisted DBMs surpass the performance of significantly larger standard DBMs, centered DBMs and eclipse the performance of much larger pre-trained DBMs. In fact, with a DBM of only $120\times 18$ hidden nodes (and no pre-training), trained with our mode-assisted approach, we reach the same log-likelihood of a DBM with $500 \times 1000$ hidden nodes {\it and} pre-training -- a parameter savings of two orders of magnitude.

For a fair comparison, the total training iterations were set to 100 epochs in each case, and the log-likelihood results with our approach are shown as a function of total number of hidden nodes at a fixed ratio, $\alpha = 0.15$. The partition function was approximated using Annealed Importance Sampling (AIS) (see Methods), identical to the other referenced results~\cite{melchior2016center}. It's worth noting that AIS tends to {\it under}-approximate the partition function, leading to an over-estimate of the log-likelihood. This effect is exaggerated in larger models. Even with this negative impact, a mode-assisted DBM with a total of 138 nodes without pre-training achieves about the same performance as one with 1,500 that was pre-trained.

\section*{Discussion}
We conclude by providing an intuitive understanding of the dramatic performance improvement of the mode-assisted training over Gibbs sampling (CD). The standard method of pretraining DBMs initializes weights at a greedy starting point but critically ignores correlations within the system. The end result is a dramatic loss of parameter efficiency compared to mode-assisted joint training of DBMs. Even training methods like centering, with more advanced sampling techniques like parallel tempering~\cite{swendsen1986replica}, fail to train the DBM near its capacity. The issue is the reliance on Gibbs-like sampling to explore the states of the DBM, and using only these statistics to compute the likelihood gradient. The breakdown occurs when Gibbs chains fail to explore adequately the states of the system. 

Because of its random-walk nature, the local dynamics of Gibbs sampling suffers long auto-correlation times when equally likely states are separated by an extensive number of variables flips. As a consequence of long mixing times between such states, statistics become heavily biased. In the initial phase of DBM training, when weights are small and randomly distributed, these issues are minimized. Interactions between nodes in this `high temperature' regime are weak, and are well approximated by a mean-field theory~\cite{sherrington1975solvable}. In this phase, CD works well without much bias, and is the reason mode-assistance is really not necessary early on in the training.

However, as training continues, the DBM is effectively `cooled' as weights learn from the data and grow larger in magnitude. Correlations between nodes become significant, and can no longer be ignored or averaged over: mean-field theory is inadequate to describe this phase. Gibbs sampling can then dramatically fail in these conditions, that is why mode-assistance is primarily applied in this phase. During mode updates, information is propagated from the ground state, the most `collective' or `coordinated' state of the DBM, to all the weights. This prevents probability mass density from accumulating far away from the current state of the chain, keeping Gibbs sampling in its effective regime.

Aside from mode-assistance, we have discovered that the topology of a DBM plays a major role in its performance as well as its parameter efficiency. Too many hidden nodes act as a noise source during training, slowing down learning. Too few hidden nodes reduce the capacity of the network. For the network sizes considered, a `sweet spot' was found where these two effects are in balance. At this stage, it is not clear if this is particular to DBMs, or there are deeper reasons why this is the case, and further work in this direction would be interesting. 

Finally, mode training relies on an efficient search for the mode of a DBM, which is an NP-hard computational problem on its own. On this front, we employ a novel dynamical systems based optimization technique called memcomputing, which has shown great promise in efficient optimization of non-convex energy landscapes like the DBM~\cite{sheldon2019taming,manukian2020mode}. Since the DBM energy landscape can be represented within the optimizing dynamics, a path exists to extend these systems to (learning) samplers. In doing so, the need for CD 
would be entirely eliminated. We leave these avenues of research to be explored in future work.

\section*{Methods}
DBMs are powerful models in machine learning, but bring with them considerable computational burdens during evaluation. To deal with them, we follow the procedure outlined in Ref.~\onlinecite{salakhutdinov2009deep}. First, due to additional complexity, the log-likelihood is not directly maximized as in the case of RBMs. Instead, a variational lower bound to the log-likelihood is optimized. Second, the most common variational form chosen leads to mean-field updates for the data term. Finally, for evaluation of log-likelihood lower bounds, an approximation of the partition function is necessary. We provide a short description of all these approximations which are frequently encountered in DBM training scenarios.

{\it Likelihood Lower Bound} - Since the data expectation in Eq.~(\ref{eq:w_gradient}) is no longer in closed form for the DBM, data-dependent statistics must be approximated with a sampling technique over the conditional distribution, $p({\bf h} | {\bf v})$, where ${\bf h} = \{ {\bf h}^{(1)}, \cdots, {\bf h}^{(\ell)}\}$. In practice, a variational lower bound to the log-likelihood is maximized instead, which is tractable and is found to work well (as in the model term)~\cite{neal1998view,salakhutdinov2009deep,salakhutdinov2012efficient}.

The variational approximation replaces the original posterior distribution, $p({\bf h} | {\bf v})$ with an approximate distribution, $r({\bf h}| {\bf v})$, where the parameters of $r$ follow the gradient of the resulting {\it lower bound} on the original log-likelihood~\cite{neal1998view}, 

\begin{align}
\label{eq:elbo}
\log p({\bf v}) &\geq \sum_{{\bf h}} r({\bf h} | {\bf v}) \log p({\bf v}, {\bf h}) + H_r({\bf v})\\
&= \log p({\bf v}) - \textrm{KL}(r({\bf h}|{\bf v}) || p({\bf h}| {\bf v})) \nonumber,
\end{align}
where $H_r({\bf v})=-\sum_{\bf h}r({\bf h} | {\bf v})\log r({\bf h} | {\bf v})$ is the Shannon entropy of $r$. This variational loss simultaneously attempts to maximize the log-likelihood of the data set and minimize the KL divergence between the true conditional distribution, $p$ and its approximation, $r$.

{\it Data Term} - A fully factorial mean field ansatz is often used in the variational approach, $r({\bf h}|{\bf v}) = \prod_i r(h_i | {\bf v})$ with $r(h_i = 1| {\bf v}) = \mu_i$, with $\mu_i\in [0,1]$ randomly initialized from a uniform distribution. Maximizing the lower bound in  Eq.~(\ref{eq:elbo}) results in updates of the form,
\begin{equation}
\mu_i \leftarrow \sigma\left(\sum_jW_{ij}^{(1)}v_j + \sum_{i\neq j} J_{ij}\mu_i + b_i\right).
\end{equation}
Here $J$ is a block matrix containing the weights of the hidden nodes and $b_i$ is the bias of the $i$-th hidden node. Convergence is typically fast, (in our experiments less than 30 iterations are enough) and these states are then used to calculate the data expectation in Eq.~(\ref{eq:w_gradient}) during the Gibbs phase of the training. 

{\it Partition Function} - Computing the partition function in Eq.~(\ref{eq:pvh}) exactly is infeasible for large DBMs. Its value is only required to evaluate the performance of the networks and appears in the log-likelihood as a normalizing constant. Annealed Importance Sampling (AIS)~\cite{salakhutdinov2008learning} is the procedure often used to approximate the partition function of large RBMs and DBMs. AIS estimates the ratio of partition functions, $Z_N/Z_0$, using a sequence of probability distributions between a chosen initial distribution and the desired one. The initial distribution, $p_0$, is chosen to have an exactly known partition function and to be simple to sample from (e.g. uniform) and $p_N$ is the desired distribution whose partition function one wants to compute. 

The sequence in the case of DBMs is parameterized by $\beta_k$ (inverse temperatures), giving $p_k({\bf x}) = e^{-\beta_k E({\bf x})}/Z_k$. Markov chains are initialized uniformly according to $p_0$ and $0 \leq \beta_k \leq 1$ is slowly annealed to unity according to a desired schedule, all the while allowing the chains to run. The ratio is then approximated by the product of ratios of the unnormalized intermediate distributions,

\begin{equation}
\frac{Z_{N}}{Z_0}= \frac{p_1^*({\bf x})}{p_0^*({\bf x})} \cdots \frac{p_k^*({\bf x})}{p_{k-1}^*({\bf x})}.
\end{equation}

When dealing with bipartite graphs like DBMs, either all the even or all the odd layers can be analytically traced out, resulting in a tighter approximation. For a direct comparison with previous work~\cite{salakhutdinov2009deep,melchior2016center}, we average over an identical number of 100 AIS runs with 29,000 linearly spaced intermediate distributions.

\noindent {\bf Acknowledgments}\\
Work supported by DARPA under grant No. HR00111990069. H.M. acknowledges support from a DoD-SMART fellowship. 
All memcomputing simulations reported in this paper have been done on a single core of an AMD EPYC server. \\

\bibliographystyle{naturemag}
\bibliography{../refs}

\end{document}